\documentclass[letterpaper, 10 pt, conference]{ieeeconf}  

\IEEEoverridecommandlockouts                              

\usepackage{amsmath} 
\usepackage{amssymb}  
\usepackage{graphicx}  
\usepackage{tabularx}
\usepackage{booktabs}
\usepackage{color}
\usepackage{textgreek}

\usepackage{algorithm}
\usepackage{algorithmic}
\usepackage{subcaption}

\title{\LARGE \bf
Substrate-Timing-Independence for Meta-State Stability of Distributed Robotic Swarms
}

\author{Tinapat Limsila, Mehul Sharma, and Paulo Garcia
\thanks{International School of Engineering, Faculty of Engineering, Chulalongkorn University, Thailand
        {\tt\small paulo.g@chula.ac.th}}%
}

\begin{document}

\maketitle
\thispagestyle{empty}
\pagestyle{empty}

\begin{abstract}

Emergent properties in distributed systems arise due to timing unpredictability; asynchronous state evolution within each sub-system may lead the macro-system to faulty meta-states. Empirical validation of correctness is often prohibitively expensive, as the size of the state-space is too large to be tractable. In robotic swarms this problem is exacerbated, when compared to software systems, by the variability of the implementation substrate across the design, or even the deployment, process.
\par We present an approach for formally reasoning about the correctness of robotic swarm design in a substrate-timing-independent way. By leveraging concurrent process calculi (namely, Communicating Sequential Processes), we introduce a methodology that can automatically identify possible causes of faulty meta-states and correct such designs such that meta-states are consistently stable, even in the presence of timing variability due to substrate changes. We evaluate this approach on a robotic swarm with a clearly identified fault, realized in both simulation and reality. Results support the research hypothesis, showing that the swarm reaches an illegal meta-state before the correction is applied, but behaves consistently correctly after the correction. Our techniques are transferable across different design methodologies, contributing to the toolbox of formal methods for roboticists.

\end{abstract}

\section{INTRODUCTION}

The advent of fully-distributed robotic swarms promises improvement on diverse applications where human intervention is not desirable and reliability is paramount \cite{chung2018survey}. In the absence of a single point of failure, self-organizing, autonomous swarms can intelligently adapt themselves to the situation at hand and perform operations beyond the scope or reach of a single robot \cite{soares2018group}.

\begin{figure}[t]
    \centering
    \includegraphics[width=0.7\columnwidth]{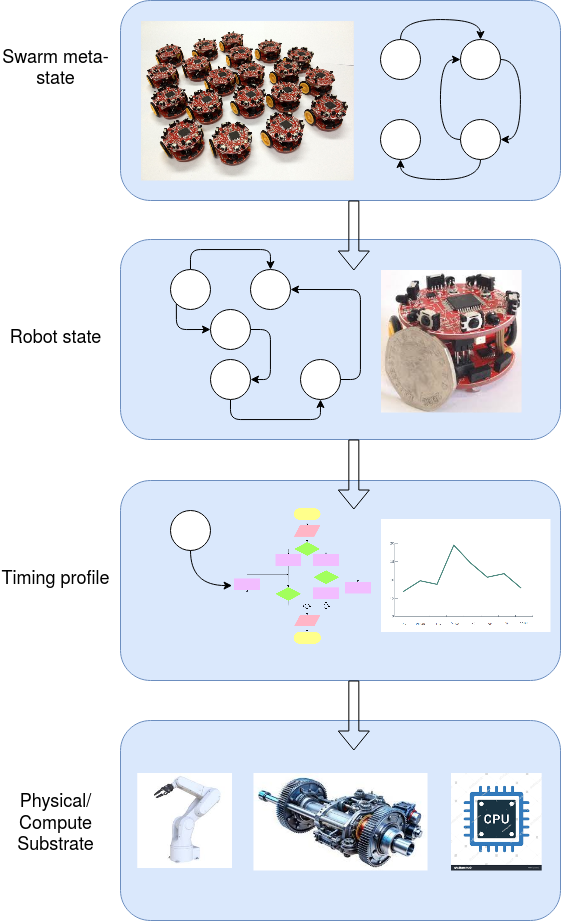}
    \caption{Mapping from collective meta-state to individual robot state, as a function of timing profiles of the underlying physical and compute substrate. Parts of the image replicated from third-parties under a Creative Commons License.}
    \label{fig:substrate}
\end{figure}

\par Efficient design and deployment of robotic swarms is fraught with challenges: from kinematic considerations on multi-robot actuator systems \cite{dimarogonas2008connectedness}, to distributed path planning or localization and mapping \cite{zhong2023dcl}. Our focus is on the challenge of meta-state stability, which can be formulated as: "\textit{Given a swarm of $k$ robots, where each is in a certain logical state, how to ensure the swarm's meta-state is logically consistent and valid?}". I.e., how to ensure asynchronous, local state to state transitions do not lead the system to a faulty (e.g., deadlocked, livelocked) meta-state? We further specialize this research question to focus on solutions that are substrate-timing-invariant \cite{vasiljevic2021compute}, where substrate refers to the specific technology and/or components utilized to effect each robot (including different choices of sub-systems, from processing to motion, to different test media such as physical prototypes or digital simulation \cite{park2020flexible}), and timing refers to the required time, and associated variation, for each state to state transition to occur as a function of substrate properties \cite{yasuda2021mechanical}.

\par In other words, we are concerned with solutions that can be reliably transferred when substrate changes. This is particularly relevant with the software-ization of robotic design, where different incarnations of a prototype evolve over successive simulations, of increasing level of complexity, until physical prototypes are realized; and, with the component-ization of robotics, leading to future designs that may hot-swap sub-systems in response to non-functional requirements \cite{garcia2024preserving}. Thus, our focus is on \textit{correct by design} solutions that are agnostic to the temporal behavior offered by the underlying substrate \cite{ducatelle2011self}. To this end, we present and evaluate a methodology leveraging \textit{Communicating Sequential Processes} for substrate-timing-independence of robotic swarms.  Specifically, this paper offers the following contributions:

\begin{itemize}
    \item With regards to swarm robotics, we formalize the process of transferring designs across different substrates, identifying critical aspects that contribute to temporal mismatch by introducing intra- and inter-robot timing variations that lead to emergent phenomena.
    \item We describe a process for reliable design that guarantees prevention of such emergent phenomena by leveraging Communicating Sequential Processes (CSP) as a calculus for inter-robot behavioral description and analysis.
    \item We show such a process leads to high-level behavior that is invariant in regards to intra- and inter-robot timing, ensuring reliable transfer across substrates.
    \item We perform an empirical evaluation of such process on a fully-distributed swarm system.
\end{itemize}

\section{BACKGROUND}

The effects of choice of computing substrate are well known in the computer engineering world: e.g., Graphical Processing Units (GPUs) typically offer much higher performance than general purpose processors for embarrassingly parallel workloads, at the cost of determinism; execution-time bounds are generally harder to define, and typically larger, making them unfeasible for safety-critical real-time applications \cite{perez2022gpu}. In robotics, substrate choice has typically focused on functional requirements (e.g., choice of grippers, locomotion possibilities), but its impact on non-functional properties is under-explored. Whilst this has negligible impact in the case of a single robot, it has serious consequences to the design of robotic swarms, as the (generally negative) emergent properties of distributed systems arise from this choice, if not properly handled.

\par The challenges of distributed computation emerge from the asynchronous nature of processing (both digital and physical) across distinct systems (either robotic or software). As we can not offer any \textit{a priori} guarantee of timing, especially regarding communication latency, it is possible that specific components perceive effect before cause; a classic result in distributed systems, first shown by Lamport in his classic paper \cite{lamport2019time}. Because of this, logical sequences of events (from an independent observer's frame of reference) may be perceived by different components in different order, leading to local state transitions that correspond to global faulty meta-states, i.e., heisenbugs \cite{ramesh2025unveiling} (we elaborate on and formalize this aspect in Section III). Whilst these issues can (to a certain extent) be identified and mitigated on a fixed substrate, \textit{ad hoc} solutions inevitable fail when the underlying substrate changes, as potentially all previously profiled timing characteristics change accordingly \cite{hosking2010testability}.

\par In distributed software, particularly in the sub-domain of formal systems and certification, it is commonplace to model systems using some formal process calculus that allows one to reason about distributed heisenbugs in a substrate-timing-independent way, by lifting the distributed design process to a higher level of abstraction; i.e., such that its reduction rules correspond to timing-independent actions in the domain. Examples of such calculi include formal specifications of Petri Nets \cite{abdul2021petri}, Communicating Sequential Processes (CSPs) \cite{hoare2021communicating}, the \textpi-calculus \cite{rounds2004spatial}, among others. We will use CSP throughout the rest of this paper, but our approach is not specific to this particular calculus.

CPS contains two primitives: \textit{events} (notated in lower case) and \textit{processes} (notated in upper case), which include the fundamental instances \textit{SKIP} (terminate successfully) and \textit{STOP} (terminate with fault). A process is defined in function of sequences of events and other processes, under the rules and corresponding notation given by: $\mathbf{P: a \rightarrow Q}$ (P admits event \textit{a} then behaves as Q); $\mathbf{P: Q;R}$ (P is the sequential composition of processes Q and R); $\mathbf{P: (a \rightarrow Q) \Box (b \rightarrow R)}$ (P admits an external choice of either event \textit{a} or \textit{b} then behaves as Q or R); $\mathbf{P: (a \rightarrow Q) \sqcap (b \rightarrow R)}$ (P internally chooses either event \textit{a} or \textit{b} then behaves as Q or R); and $\mathbf{P: Q ||| R}$ (P is the independent, concurrent executions of processes Q and R). Processes are further composed by the operation $\mathbf{Q|[e_1,e_2,...,e_n]|R}$ (processes Q and R operate in parallel, but synchronized on events $\{e_1:e_n\}$); and can communicate with each other through named channels, such that $c!m$ is the event that writes message $m$ to channel $c$, and $c?m$ is the event that reads message $m$ from channel $c$. Some authors and software versions further allow the use of algebra on messages, such that $c_o!(2 \times c_i?m)$ outputs $m\times2$ on channel $c_o$, where $m$ is read from channel $c_i$.

\section{PROBLEM STATEMENT}

Each robot $R_i$ in a swarm is functionally specified by a finite state machine, denoted  $\text{fsm}_i$. Each finite state machine is specified by the tuple $(\sum_i, S_i, s_i^0 \in S_i, \delta_i)$, where $\sum_i$ is the input alphabet (i.e., the finite set of internal and external events that trigger state machine evolution), $S_i$ is the set of all possible robot states,  $s_i^0 \in S_i$ is the initial state, and $\delta_i : (S_i,\sum_i) \rightarrow S_i$ is the state transition function, given current state and input symbols. For simplicity, we assume that $S_i$ includes only functionally correct states (e.g., states only reachable through hardware faults are not considered) and $\delta_i$ is a deterministic function across its entire domain. Throughout the remainder of this paper, we further assume that, for a swarm of $k$ robots, $\text{fsm}_i = \text{fsm}_j, \forall i,j \in [0:k-1]$; i.e., we assume swarms of homogeneous robots for simplicity, although our results can be extended for heterogeneous swarms.

The swarm's global meta-state $s_{sw} \in S_{sw}$ is given by the combinations of each robot's state at time $t$, denoted by $s_{sw}^t = \{s_0^t,s_1^t,...,s_{k-1}^t\}$, such that $|S_{sw}| = |S|^k$, for a swarm of $k$ robots. The problem of state inconsistency arises from the fact that several states in $S_{sw}$ are illegal states, i.e., correspond to a faulty combination of robot states (inconsistency). Identifying such illegal states, and whether or not they are reachable, is non-trivial, as there is no explicit swarm state transition function $\delta_{sw} : (S_{sw},\sum_{[0:k-1]}) \rightarrow S_{sw}$: this function emerges from asynchronous firings of each robot's transition function $\delta_i$. This problem is well established in the distributed systems community, where several computing solutions (for strictly software systems) have emerged over the years: we point interested readers to the \cite{ahmed2013survey} for a comprehensive review.

\begin{figure*}[t!]
    \centering
    \includegraphics[width=0.7\textwidth]{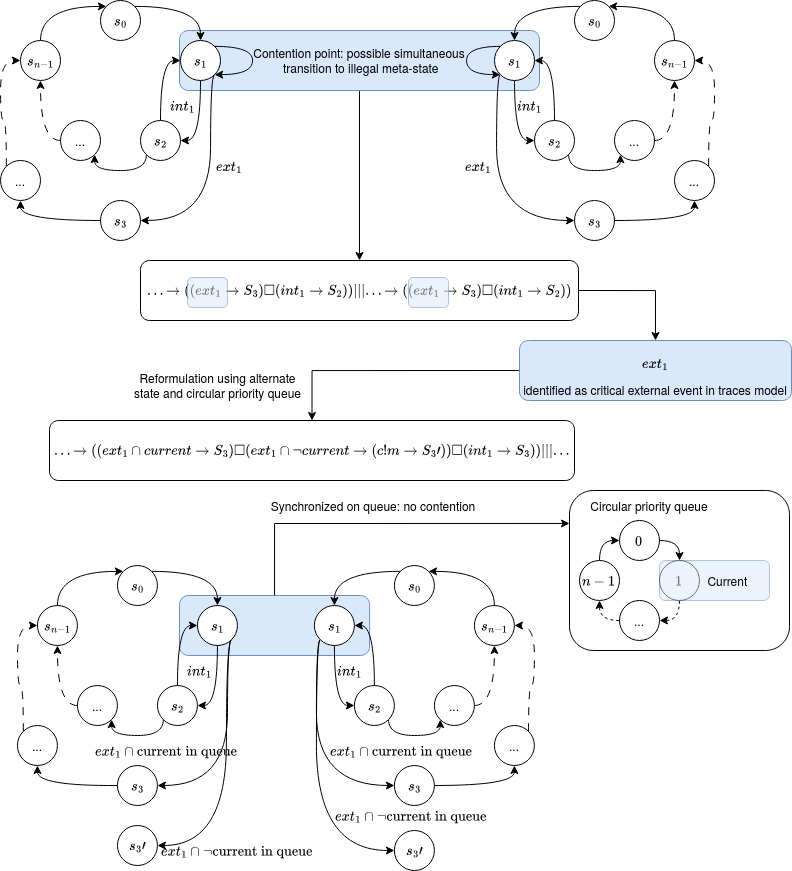}
    \caption{Design flow for identification of possible illegal meta-state transitions and corresponding re-design, synchronized on globally consistent circular priority queue.}
    \label{fig:csp-flow}
\end{figure*}

\par In swarm robotics, the problem is further exacerbated by two domain aspects. First, solutions relying on shared or centralized state are not applicable to fully-distributed swarm robotics, as any one robot acting as a central coordinator becomes a single point of failure, and establishing fault-tolerance \cite{bjerknes2013fault} techniques further increases the number of state combinations. Second, whilst in software systems design, testing, and deployment occur in the same medium (i.e., there exists a negligible difference between conceptual "simulation" and "deployment"), and several properties remain invariant across testing and deployment, the same is not true in the robotics domain. Simulation conditions, useful for quick prototyping, differ from reality in potentially drastic ways \cite{ligot2020simulation}.

\par We focus on timing variation caused by substrate changes between realizations: e.g., time required for localization, sensing, actuator control, movement, etc. Ensuring that a swarm does not reach an illegal meta-state in one substrate (digital or physical), through extensive testing, does not guarantee the same reliability in another deployment, when the time between each robot's state changes, compared to the prior version, leading to asynchronous combinations of individual state transition functions that were not previously seen. Thus, correctness by design, rather than empirical verification, is the preferred approach to swarm robotics design.

\par For each robot $R_i$, time between each successive state can be modeled as a state transition matrix $\mathbf{T}_i$:

\begin{equation}
    \mathbf{T}_i = \begin{bmatrix}
                    t_{0,0} & t_{0,1} & t_{0,2} & \dots  & t_{0,n-1} \\
                    t_{1,0} & t_{1,1} & t_{1,2} & \dots  & t_{1,n-1} \\
                    \vdots & \vdots & \vdots & \ddots & \vdots \\
                    t_{n-1,0} & t_{n-1,1} & t_{n-12} & \dots  & t_{n-1,n-1} 
                 \end{bmatrix}
\end{equation}

where $n$ is the number of states, and $t_{i,j}$ denotes the time required for a transition from state $i$ to state $j$ (including permanence in state $i$), expressed as $t_{i,j} = \bar{t_{i,j}} \pm \Delta t_{i,j}$. If a function $\delta(s_i,\sum)$ admits no image $j$, $\forall \epsilon \in \sum$, then $t_{i,j} = \infty$; if it admits, it is treated as an unknown to model unpredictable timing. Given a set of state orbits $\Pi = \{\pi_0,\pi_1,...,\pi_{k-1}\}$, where each orbit corresponds to a robot's state evolution over time, i.e, $\pi = [s_{t0:t1} \in S,s_{t1:t2} \in S,...,s_{ta:tb} \in S]$, where times $\{ta:tb\}, \forall a,b \in \mathbb{R}$ are sampled from $\mathbf{T}$ for each specific state transition, the problem can thus be formulated as: how to specify the sets $S_i$ and the state transition functions $\delta_i : (S_i,\sum_i) \rightarrow S_i$, such that  $\forall t \in \mathbb{R}, \nexists$ $ \Pi \rightarrow \{s_0^t,s_1^t,...,s_{k-1}^t,\} \in \underline{S_{sw}}$, where $\underline{S_{sw}} \subset S_{sw}$ denotes the set of illegal states in $S_{sw}$.

\par In a statistical approach, each orbit $\pi$ could be treated as a Markov Chain obtained from $fsm_i$ and $\mathbf{T}$, and Monte-Carlo methods could be applied to obtain reachability guarantees (e.g., \cite{deshmukh2018mean}). In contrast, we focus on using formal methods to guarantee correctness-by-design; i.e., using Communicating Sequential Processes (CSP) to guide the design of $S_i$ and $\delta_i : (S_i,\sum_i) \rightarrow S_i$ guaranteeing no illegal meta-states can be reached.

\section{SUBSTRATE-TIMING-INVARIANT DESIGN USING CSP}

Formalizing swarm behavior in CSP automatically provides a specification that is a function of global events; i.e., events that either affect the entire swarm or are the result of intra-swarm communication, as all internal (to each robot) events are abstracted by the CSP view of a process. In essence, a CSP description of a robot's process, denoted $csp_i$, corresponds to a simplified state machine that considers only externally visible events (internal events and transitions are abstracted into sub-state-machines that make up each state in the CSP view). We then leverage CSP's \textit{traces model} \cite{cavalcanti2020inputs}, well established in the literature, to identify deadlock and livelock conditions in a swarm specification \cite{lima2020framework}.
\par Once triggering events are identified, we modify relevant state transitions to include a \textit{consensus} negotiation based on a globally synchronized priority queue. The algorithm is described in Alg. 1, and an example is depicted in Fig. \ref{fig:csp-flow}.

\begin{algorithm}
\begin{algorithmic}
\STATE \textbf{Input:} $fsm = (\sum, S, s^0 \in S, \delta)$, $csp$, number of robots $k$
\STATE \textbf{Output:} $fsm\prime = (\sum, S\prime, s^0 \in S\prime, \delta\prime)$,  priority queue
\STATE queue $: createQueue(k)$
\STATE $fsm\prime : fsm$
\STATE swarm  $: csp_0 ||| csp_1 ||| ... ||| csp_{k-1}$
\FORALL{$\bar{t_{i,j}} + \Delta t_{i,j} \in \mathbf{T}$}
    \FORALL{$\bar{t_{i,j}} - \Delta t_{i,j} \in \mathbf{T}$}
    \STATE trace = $GenerateTrace(\pi(swarm))$ 
    \IF{$locked(csp\prime,trace)$}
        \STATE event $: getEvent(trace)$
        \STATE $fsm\prime : {fsm\prime | s_c}$
        \FORALL{$s,event \in D(\delta\prime)$}
            \STATE $\delta\prime(s,event) : ObtainConsensus(s,event)$
        \ENDFOR
    \ENDIF
    \ENDFOR
\ENDFOR
\RETURN{$fsm\prime$, queue}
\end{algorithmic}
\caption{CSP refactoring algorithm for swarm meta-state stability.}
\end{algorithm}

\begin{algorithm}
\begin{algorithmic}
\STATE \textbf{Input:} state/process $P$, event $e$
\STATE \textbf{Output:} $S\prime$, $\delta : (S\prime,\sum) \rightarrow S\prime$
\STATE $csp : (e \rightarrow C_m) \Box (c?s \rightarrow C_s)$
\STATE $C_m :$ \newline
    \hspace*{1em}$(m \rightarrow (c!s \rightarrow A) \sqcap (\neg q \rightarrow (c!s \rightarrow C_s)))$\newline 
        $A :$ \newline 
    \hspace*{1em}    $((c?ack_1 \rightarrow SKIP  ||| ...) ; (c!p \rightarrow P))$\newline
        $C_s :$\newline
    \hspace*{1em}     $(c_m?s \rightarrow (c!ack \rightarrow (c?p \rightarrow P))) \Box A$
\RETURN{$toFSM(csp)$}
\end{algorithmic}
\caption{"$ObtainConsensus()$" Generation of consensus states and state transition functions for a new triggering event.}
\end{algorithm}

\par It is noteworthy that, unlike distributed software systems at scale, which require some more advanced consensus algorithm such as PAXOS \cite{lamport2001paxos}, a globally synchronized priority queue suffices in the case of robotic swarms as the number is known and typically fixed throughout deployment, robots are safely assumed to be non-Byzantine, and faulty robots that temporarily leave the protocol can re-synchronize their internal view of the priority queue (although we do not depict this scenario in our description or experiments).

\section{EXPERIMENTS AND RESULTS}

\subsection{Experimental Setup}

\begin{figure}
\centering
\begin{subfigure}{.5\columnwidth}
  \centering
  \includegraphics[width=.95\linewidth]{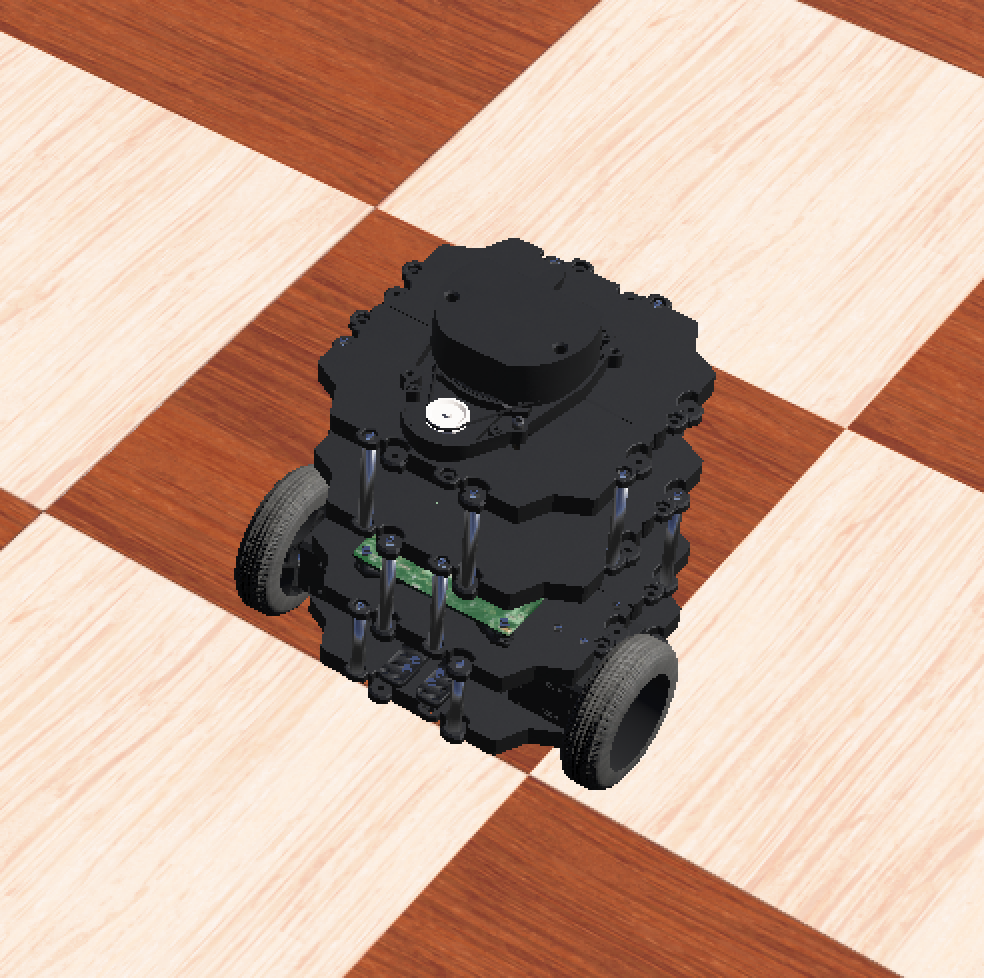}
  \caption{}
  \label{fig:sub1}
\end{subfigure}%
\begin{subfigure}{.5\columnwidth}
  \centering
  \includegraphics[width=.95\linewidth]{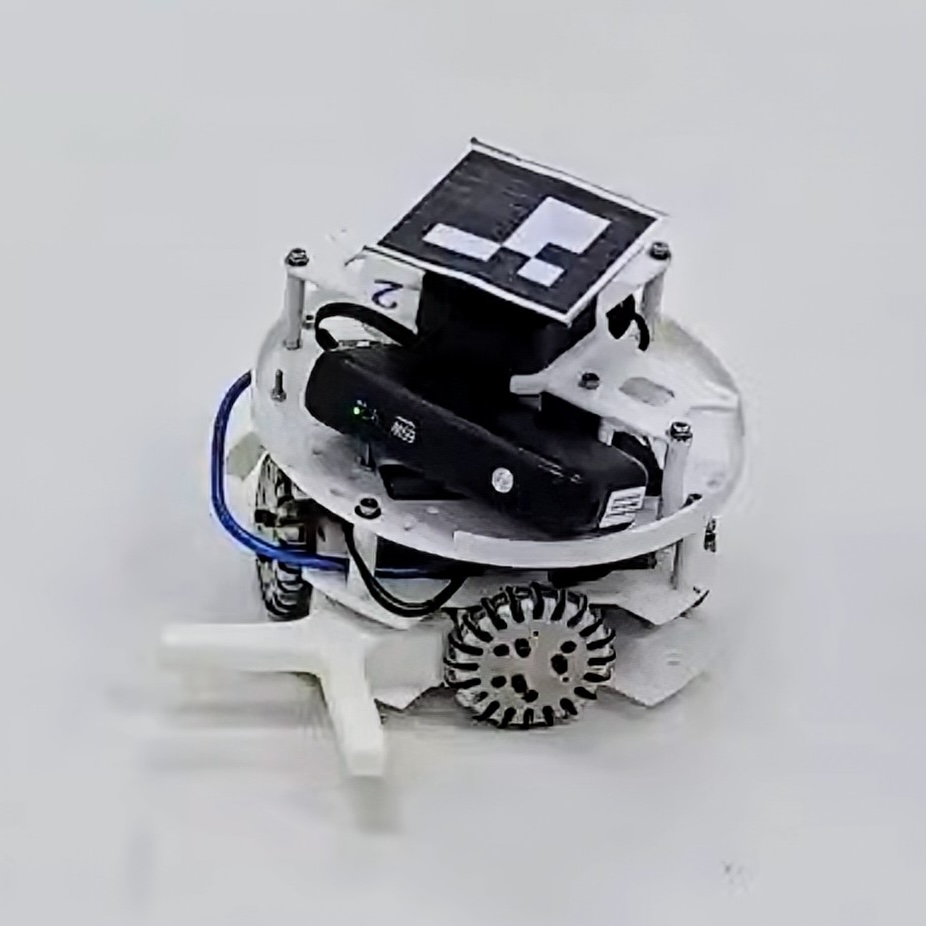}
  \caption{}
  \label{fig:sub2}
\end{subfigure}
\caption{Implemented test substrates: (a) simulation (b) physical}
\label{fig:robots}
\end{figure}

We built a robotic swarm of 3 robots in digital simulation (WeBots \cite{michel2004cyberbotics}) and an equivalent physical instance. All substrates (compute, locomotion, communication) change between instances, providing substantially different temporal behavior: specifics are described in Table \ref{table:hardware-software}, with examples depicted in Fig. \ref{fig:robots}. In both cases, the logical view of swarm behavior is the same: the swarm is meant to search and identify a specific object. Once it is found, the swarm converges on it in a symmetric approach pattern and collaboratively moves it to a pre-defined position. Robots amble in a random walk (\textit{idle} state) until at least one has detected an object. Once a potential object is located, a robot attempts to identify it (\textit{object detection} state). If it is deemed a valid target, the identifying robot determines optimal paths for each robot in the swarm (\textit{path finding} state, eliciting other robots' coordinates from direct communication) and each robot then follows its assigned path until convergence on the object, followed by transport (\textit{path following} state). The full state-machine that drives robot behavior is more complex, as each depicted state in in fact a sub-state-machine that deals with all the substrate-specific details and processing, but as all events in those sub-state-machines are purely internal, they can be abstracted away when considering the impact on swarm behavior.

\begin{table*}[t!]
    \centering
    \begin{tabular}{r | c c}
        \textbf{Parameters} & \textbf{Simulation}   & \textbf{Physical}\\
        \hline
        \textbf{Drive Train} & Dynamixel-XL430-W250 (Differential Drive)  & In-house closed-Loop DC Motor controller (Omnidirectional Drive)\\
        \textbf{Compute} & WeBots built-in program interpreter & Nvidia Orin Nano\\
        \textbf{Sensing} & WeBots built-in camera simulation & Logitech C920 Camera\\
        \textbf{Communication} & Point-to-point IR Receiver/Transceiver  & TCP/IP over WiFi
    \end{tabular}
    \caption{Hardware and compute substrates for each of the two swarm instances.}
    \label{table:hardware-software}
\end{table*}

This design contains a design flaw: it assumes only one robot identifies an object at any given time, assuming the role of path finder in a singleton fashion. If two or more robots simultaneously detect an object, the swarm enters an illegal meta-state, resulting in undefined behavior. A corrected version includes a \textit{consensus} state as per the described algorithm above, where robots who simultaneously identified an object synchronize to decide on a path-finder. In each configuration, we locally log timestamps between state transitions as non-invasively as possible, and capture data for analysis post-mortem. The CSP formulation for the flawed version is given by:

\begin{equation}
\begin{aligned}
&   R : (l \rightarrow (d \rightarrow P \sqcap \neg d \rightarrow R)) \Box (c?p \rightarrow F)\\
&   P : (f \rightarrow (c!p \rightarrow F))\\
&   F : (t \rightarrow R) 
\end{aligned}
\end{equation}

where $R$ notates a robot in idle state, $l$ notates object location, $d$ and $\neg d$ notate positive and negative detection, respectively; $P$ notates path finding and $c?p$ notates reception of a path to follow from a given communication channel and $c!p$ notates equivalent transmission of a path, $F$ notates path following, $f$ notates successful path calculation, and $t$ notates successful termination of object transportation. The corrected version is described by the following formulae:

\begin{equation}
\begin{aligned}
&   R : (l \rightarrow (d \rightarrow C_m \sqcap \neg d \rightarrow R)) \Box (c?s \rightarrow C_s)\\
&   C_m : (m \rightarrow (c!s \rightarrow A) \sqcap (\neg q \rightarrow (c!s \rightarrow C_s)))\\ 
&   A : ((c?ack_1 \rightarrow SKIP ||| c?ack_2 \rightarrow SKIP ||| ...) ; (c!p \rightarrow P))\\
&   C_s : (c_m?s \rightarrow (c!ack \rightarrow (c?p \rightarrow P))) \Box A\\
&   P : (f \rightarrow (c!p \rightarrow F))\\
&   F : (t \rightarrow R) 
\end{aligned}
\end{equation}

where $m$ notates being at the top of the priority queue (master), $C_m$ notates consensus seeking when a robot is at the top of the priority queue, $c!s$ notates sending a synchronization message to all other robots, $c?ack$ and $c!ack$ notate receiving and sending an acknowledgement, respectively, $A$ notates waiting for all acknowledgements, $C_s$ notates consensus seeking when a robot is not the master, and $c_m?s$ notates receiving a synchronization request from a robot with higher priority.





\subsection{Results and Discussion}

Swarm profiling yielded the results depicted in Table \ref{tab:results}. As predicted by the hypothesis, both simulation and physical realization of the faulty swarm result in faulty meta-states, but their incidence is higher in physical realization: this is not surprising, as the timing variation offered by the substrate is of the same order of magnitude as the average time (in contrast, simulation substrate exhibits variation of smaller orders of magnitude, so there are fewer instances in the execution state-space that lead to faulty meta-states).
After applying the correction algorithm and implementing the corresponding transformation for establishing consensus, we observed no instance of faulty meta-state across either substrate, supporting the validity of the formal approach through empirical verification.
Our presentation and experimentation focused on a single event, common across robots, that can lead to faulty meta-states: it is possible, in more complex state machines, that only specific sequences of multiple events lead to faults. In such case, our algorithm would still detect the first event that potentially triggers faults, and consensus establishment would be applied there. From a performance perspective, this leads to degradation because of pessimism; a more elegant solution would only establish consensus at the processing of the last event in a chain that leads to faults, if required, but such a technique is left to future work.

\begin{table}
\centering
\begin{tabular}{c | c | c}
$\mathbf{s_i \rightarrow s_k}$ & \textbf{Simulation} & \textbf{Physical}\\
\hline
$s_0 \rightarrow s_1$ & $0.07 \pm 0.02$ & $1588.60 \pm 1390.80$\\
$s_1 \rightarrow s_2$ & $160.19 \pm 4.50$ & $98.90 \pm 57.70$\\
$s_2 \rightarrow s_3$ & $31.16 \pm 1.66$ & $116.10 \pm 56.50$\\
$s_0 \rightarrow s_4$ & $129.04 \pm 2.99$ & $1698.10 \pm 1414.00$\\
$s_4 \rightarrow s_3$ & $37.48 \pm 1.90$ & $79.00 \pm 26.70$
\end{tabular}
\caption{Timing results (mili-seconds) for state to state transition across different substrates. $s_0$: object detection; $s_1$: consensus; $s_2$: path finding; $s_3$: path following; $s_4$: idle. Times from $s_3 \rightarrow s_4$, $s_4 \rightarrow s_0$ and $s_4 \rightarrow s_3$ are environment-specific, not depicted.}
\label{tab:results}
\end{table}

\section{RELATED WORK}

\par The use of simulation as a way to identify emergent phenomena in robotic swarms is a well-established practice \cite{azarnasab2007integrated}. Addressing the impact of substrate changes \cite{ligot2020simulation}, however, has mostly focused on intra-robot properties, e.g., performance of individual sub-systems. Regarding inter-robot emergent phenomena, most related work has focused on testing practices, often supported by specific frameworks: e.g., the RoboChart Domain Specific Language for modelling and verification of heterogeneous collections of interacting robots \cite{cavalcanti2018modelling} or Discrete Event System Specification for agent-in-the-loop simulation and coverage of large swarm state-spaces \cite{hosking2010testability}. In contrast, our work eschews extensive testing in favor of  the use of formal calculi (CSP) for inter-robot design, akin to prior work using CSP for intra-robot verification (at the sub-system integration level) \cite{cavalcanti2019testing,kalech2006diagnosis}.

\section{CONCLUSION}

We described an approach that ensures correct-by-design distributed robotic swarm behavior. By employing the formalism of CSPs, we show it is possible to reason about meta-state stability and ensure no faulty meta-states are reached, independently of the specific temporal behavior of the underlying substrate. We described empirical evaluation of this approach on a robotic swarm implemented across different implementation substrates: digital simulation and physical realization, with variations across compute sub-systems, input sensors, communication medium, and locomotion drivers, with a clearly identified possible fault and subsequent correction.

\par Results across multiple evaluations found no instances of meta-state instability after corrections, despite significant timing variations due to substrate variation, supporting the hypothesis that this formal approach suffices for substrate-timing-independence. Our approach is complementary, rather than alternative, to other robotic swarm design techniques, be them formal or empirical, and contributes to the formal body of knowledge in the domain, toward systematic engineering approaches that can complement and accelerate empirical testing.

\addtolength{\textheight}{-12cm}   





\bibliographystyle{IEEEtran}
\bibliography{IEEEabrv,IEEEexample}

\end{document}